\pdfoutput=1
\documentclass[11pt]{article}

\usepackage[preprint]{acl}
\usepackage{times}
\usepackage{latexsym}

\usepackage[T1]{fontenc}
\usepackage[utf8]{inputenc}

\usepackage{microtype}

\usepackage{inconsolata}

\usepackage{amsthm}
\usepackage{amsmath}
\usepackage{amssymb}
\usepackage{caption}
\usepackage{subcaption}
\usepackage{enumerate}
\usepackage{multirow}
\usepackage{graphicx} 
\usepackage{url}
\usepackage{booktabs}
\usepackage{tabularx}
\usepackage{colortbl}
\usepackage{algorithm}
\usepackage{algorithmic}
\usepackage{makecell}
\usepackage{tcolorbox}
\usepackage{enumitem}
\usepackage{xcolor}

\definecolor{color1}{HTML}{D0E2BF}  %D0E2BF
\definecolor{color2}{HTML}{d0d5e8}   %F7E6D8 c2bff7 dadcf9
\title{Activating Distributed Visual Region within LLMs for \\
Efficient and Effective Vision-Language Training and Inference}

\author{Siyuan Wang\textsuperscript{\rm 1}\footnotemark[1], Dianyi Wang\textsuperscript{\rm 2,3}\thanks{~Equal contribution.},
Chengxing Zhou\textsuperscript{\rm 4}\footnotemark[1], \\
\textbf{Zejun Li}\textsuperscript{\rm 2},
\textbf{Zhihao Fan}\textsuperscript{\rm 5}, \textbf{Xuanjing  Huang}\textsuperscript{\rm 2} \textbf{Zhongyu Wei}\textsuperscript{\rm 2,3} \\
\textsuperscript{\rm 1}University of Southern California,
\textsuperscript{\rm 2}Fudan University,\\
\textsuperscript{\rm 3}Shanghai Innovation Institute, 
\textsuperscript{\rm 4}Sun Yat-sen University,
\textsuperscript{\rm 5}Alibaba Inc. \\
\texttt{sw\_641@usc.edu; dywang24@m.fudan.edu.cn; zhouchx33@mail2.sysu.edu.cn} \\
}

\begin{document}
\maketitle
\begin{abstract}
Large Vision-Language Models (LVLMs) typically learn visual capacity through visual instruction tuning, involving updates to both a projector and their LLM backbones. Inspired by the concept of a visual region in the human brain, we investigate the existence of an analogous \textit{visual region} within LLMs that functions as a cognitive core, and explore the potential of efficient training of LVLMs via selective layers tuning. Using Bunny-Llama-3-8B-V for detailed analysis and other three LVLMs for validation across diverse visual and textual tasks, we find that selectively updating 25\% of LLMs layers, when sparsely and uniformly distributed, can preserve nearly 99\% of visual performance and maintain or improve textual task results, while effectively reducing training time. 
Based on this targeted training approach, we further propose a novel visual region-based pruning paradigm, removing non-critical layers outside the visual region, which can achieve minimal performance loss. This study offers an effective and efficient strategy for LVLM training and inference by activating a layer-wise visual region within LLMs, which proves consistently effective across different models.
\end{abstract}

\section{Introduction}
Large Vision-Language Models (LVLMs)~\cite{li2023blip,zhu2023minigpt,bai2023qwen,liu2024visual} have emerged as an increasing research interest for interpreting and interacting with the world through both visual and linguistic channels. Existing LVLMs generally utilize advanced Large Language Models (LLMs), like FlanT5~\cite{chung2022scaling} and Vicuna~\cite{vicuna2023}, as their cognitive core, and align visual features from visual encoders with LLMs' knowledge and reasoning abilities. This alignment has demonstrated remarkable performance across diverse visual tasks~\cite{lu2022learn, liu2023mmbench, fu2024mme}.

\begin{figure*}[h]
    \centering
    % 设置两个 minipage 环境，使图片和表格并列
    \begin{minipage}{0.51\textwidth}
        \centering
        \vspace{-3mm}
        \resizebox{1.0\textwidth}{!}{
        \setlength{\tabcolsep}{3.5pt}
        \renewcommand{\arraystretch}{1.1}
        \begin{tabular}{c|>{\columncolor{color1}}c>{\columncolor{color1}}c|>{\columncolor{color2}}c>{\columncolor{color2}}c}
        \toprule
        \multirow{2}{*}{Model Variants} & \multicolumn{2}{>{\columncolor{color1}}c|}{Visual} & \multicolumn{2}{>{\columncolor{color2}}c}{Textual} \\
            & OCRVQA & DocVQA & WikiText & Pile-10k \\
           \midrule
           LLaVA & 2.43 & 30.55 & 11.44 & 29.58 \\
           \midrule
           LLaVA$_\textbf{r}$ (layer 0$\sim$7) & 1.87 & 38.49 \textbf{[$\uparrow$]} & 11.37 \textbf{[$\uparrow$]} & 29.19 \textbf{[$\uparrow$]} \\
           LLaVA$_\textbf{r}$ (layer 8$\sim$15) & 1.93 & 32.35 \textbf{[$\uparrow$]} & 11.38 \textbf{[$\uparrow$]} & 29.21 \textbf{[$\uparrow$]} \\
           LLaVA$_\textbf{r}$ (layer 16$\sim$23) & 2.18 & 16.47 & 11.35 \textbf{[$\uparrow$]} & 29.33 \textbf{[$\uparrow$]} \\
           LLaVA$_\textbf{r}$ (layer 24$\sim$31) & 2.11 & 17.47 & 11.36 \textbf{[$\uparrow$]} & 29.27 \textbf{[$\uparrow$]} \\
           \midrule
           Vicuna (all layers) & 80.75 & 175.10 & 11.32 & 28.38 \\
        \bottomrule
        \end{tabular}
        }
        % \captionof{table}{Perplexity of LLaVA-1.5-7B with selective layers (in parentheses) reverted to Vicuna parameters on visual and textual tasks. Arrows indicate perplexity increases relative to LLaVA (visual tasks) and Vicuna (textual tasks).}
    \end{minipage}
    \hfill % 添加空隙
    \begin{minipage}{0.47\textwidth}
    \centering
    \includegraphics[width=0.9\textwidth, height=4.4cm]{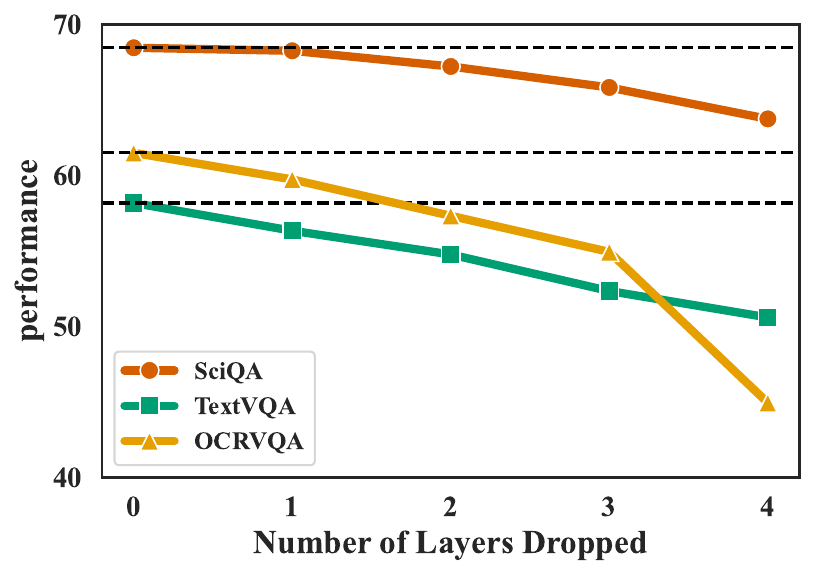}
    % \caption{Performance of LLaVA-v1.5-7B on TDIUC and MMBench while pruning certain layers according to block influence scores~\cite{men2024shortgpt}. 
    % % Results indicate that dropping a certain layers does not reduce the model's performance, suggesting the existence of layer redundancy within LVLMs.
    % }
    \end{minipage}
    \caption{\textbf{Left:} Perplexity of LLaVA with selected layers (in parentheses) reverted to Vicuna parameters on visual and textual tasks. Arrows indicate perplexity increases relative to LLaVA (visual tasks) and Vicuna (textual tasks). (1) Perplexity increases in textual tasks after multimodal training compared to the LLM backbone, indicating multimodal training compromises LLMs' linguistic abilities. (2) Perplexity decreases in visual tasks reverting certain layers (e.g., reverting layers 16–23 or 24-31 in LLaVA), suggesting these layers are redundant.
    % Both results suggest layer redundancy in LVLMs and degraded linguistic capability. 
    \textbf{Right:} Accuracy of LLaVA-1.5-7B when pruning certain layers based on angular distance scores~\cite{gromov2024unreasonable}.}
    \label{fig:layer_drop}
\end{figure*}

LVLMs are primarily trained through visual instruction tuning~\cite{liu2023visual}, which involves training both a projector and LLMs on visual instruction datasets, with optional updates to the visual encoder. Despite its efficacy, fully tuning all LLMs layers remains computationally costly, even when using efficient strategies like Low-Rank Adaptation (LoRA)~\cite{hu2021lora} and its quantized variant (QLORA)~\cite{dettmers2024qlora}. 
Additionally, extensive multimodal training risks degrading LLMs' pre-trained linguistic knowledge and reasoning capabilities~\cite{dai2024nvlm, agrawal2024pixtral}, as evidenced by LVLMs' increased perplexity on textual tasks compared to their LLM backbone in the purple section of Fig.~\ref{fig:layer_drop}.

% ~\citet{men2024shortgpt, gromov2024unreasonable} have highlighted a significant redundancy within LLMs at the layer level, enabling the removal of entire layers without substantially impacting downstream inference performance. This layer redundancy is established for LVLMs as well, as shown in Fig.~\ref{fig:layer_drop}, where LLMs serve as the core cognitive brain for visual alignment, suggesting only a portion of the LLMs may engage with visual-related tasks.
Inspired by specialized visual regions in the human brain~\cite{grill2004human} and LLMs' brain-like versatility across tasks, we propose an analogous concept of a \textit{visual region} within LLMs. We hypothesize that visual alignment to LLMs can only activate this specific \textit{visual region} while preserving LLMs' core language abilities, potentially manifesting as a layer-wise structure considering layer redundancy in LLMs~\cite{men2024shortgpt, gromov2024unreasonable}. 
We further detailedly analyze LVLMs' layer redundancy in Fig.~\ref{fig:layer_drop} (green part), shows that reverting certain layers of a LVLM to its backbone LLM' parameters minimally impacts downstream visual performance. This suggests certain layers within LLMs are non-essential for visual tasks, thereby supporting our hypothesis.
% of a layer-wise visual region for efficient LVLM training.
% Besides, after LVLM training, the LLM backbone exhibits reduced performance on linguistic tasks. All these support our hypothesis of a layer-wise visual region for efficient LVLM training.
% we demonstrate layer redundancy in LVLMs by showing that removing entire layers does not significantly affect downstream performance in Fig.~\ref{fig:layer_drop}, indicating that certain layers may be non-essential for visual tasks. We further conduct a preliminary analysis on LLaVA-1.5~\cite{liu2023improved} by reverting various layers to baseline Vicuna parameters~\cite{vicuna2023} and observe that most modifications do not significantly impact performance, thereby further supporting our hypothesis.

% Although layer-wise freezing techniques~\cite{zhang2024unified} have been developed for efficient LLM fine-tuning—where initial layers are viewed for capturing general language features and the final layers can be specialized for different tasks—these methods primarily aim to transfer to language-based tasks that rely on linguistic knowledge. This approach is not directly applicable to cross-modality transfer learning, as visual modality requires perception and alignment of visual features, which are disentangled from language representations.
% Recent task-specific parameter localization~\cite{}
% layer-wise freezing (last-layers), not suitable for multimodal
% besides, task-specific parameter localization, not generalizable
Although layer-wise freezing techniques~\cite{zhang2024unified} enable efficient LLM fine-tuning by adapting later layers for specific language tasks, they cannot be directly applied to visual tasks. Because visual alignment requires visual perception capabilities beyond textual understanding and reasoning. 
While~\citet{zhang2024overcoming} propose parameter localization for visual tasks, it remains highly task-specific and data-dependent, limiting its generalizability to versatile multimodal learning and neglecting the preservation of linguistic capabilities. To bridge this gap, we identify a general-purpose visual region within LLMs for efficient LVLM training across diverse tasks without diminishing linguistic performance. Specifically, we aim to investigate two key questions: 
(1) Where is this visual region located within LLMs?
(2) What is the necessary scale of layers in this visual region to ensure effective and efficient LVLMs training?
To this end, we embark on empirical experiments with Bunny-Llama-3-8B-V~\cite{he2024efficient} across diverse visual tasks. Our findings indicate that \textbf{sparsely and uniformly distributed layers within LLMs are the optimal position for visual learning} while simultaneously preserving textual performance. This strategic visual region selection also outperforms previous layer importance strategies. Notably, \textbf{updating only 25\% of layers achieves nearly 99\% performance on visual tasks} while effectively saving training time. We further validate this conclusion with LLaVA-1.5-7B, LLaVA-1.5-13B~\cite{liu2023visual} and Bunny-Phi3-mini-4B-V, demonstrating its consistent applicability across varying models and parameter scales. Specifically, we achieve time reductions of nearly 23\% for LLaVA-1.5-7B and LLaVA-1.5-13B, and 12\% for Bunny-Llama-3-8B-V.

Additionally, as shown in Figure~\ref{fig:layer_drop} (right), we find that commonly used layer-pruning strategies are ineffective for LVLMs, with even minimal layer removal causing significant performance degradation. In response, we propose a visual region-based pruning paradigm that selectively prunes less-important layers outside the visual region after targeted training. 
Specifically, we follow the angular distance based layer importance strategy~\cite{gromov2024unreasonable} outside the visual region,  and experimental results demonstrate that our paradigm is effective to minimizes performance decline. Overall, our work highlights promising potential for more efficient LVLMs training and inference. Notably, our approach is flexibly complementary to other efficient training techniques, such as LoRA, as demonstrated in our experiments.

% The main contributions of our paper are summarized as follows:
% \begin{itemize}[leftmargin=18pt]
%     \item We propose the concept of a specific \textit{vision region} within LLMs for more effective and efficient LVLMs training, revealing a detailed insight into functional regions within LLMs.

%     \begin{itemize}[leftmargin=18pt]
%         \item Tuning \textbf{sparsely and uniformly distributed six layers} as a \textit{visual region} can maintain or even enhance performance.
%         \item The \textit{visual region} is \textbf{consistent across in LLaVA-1.5 using various sizes of instruction-following data}, presenting a resource-efficient pathway for locating the visual region and subsequently developing LVLMs with extensive data.
%     \end{itemize}
    
%     \item We introduce a targeted training paradigm that separately trains this \textit{visual region} for perception data while using the full model for cognitive learning, even enhancing performance.
% \end{itemize}
% Main Insight/Results

\section{Preliminary of LVLMs}
\subsection{Model Architecture}
Mainstream LVLMs consist of three components: a LLM, a visual encoder, and a projector or connection module, aim to effectively leverage the capabilities of both the pre-trained visual model and LLMs. The visual encoder extracts visual features from images, commonly utilizing pre-trained models such as CLIP ViT-L/14~\cite{radford2021learning}. The connection module then projects these extracted features into word embedding space understandable by LLMs, commonly employing techniques such as linear projection ~\cite{tsimpoukelli2021multimodal}, Q-former~\cite{li2023blip}, or cross-attention layers~\cite{alayrac2022flamingo}. This enables LVLMs based on LLMs cores, like Vicuna~\cite{vicuna2023}, FlanT5~\cite{chung2022scaling}, and LLaMA~\cite{touvron2023llama} to process visual information in a similar manner as text. 

\subsection{Model Training}
The training of LVLMs can be broadly divided into two phases: pre-training and supervised fine-tuning.
Unlike LLMs, both phases utilize supervised image-text pairs for visual instruction tuning.
% \paragraph{Visual Instruction-Following Data}
% Visual instruction-following data comprise (input, output) pairs, derived from existing visual datasets covering tasks such as image captioning~\cite{chen2015microsoft, sharma2018conceptual, laioncoco, changpinyo2021conceptual}, and visual question answering (VQA)~\cite{hudson2019gqa, Singh_2019_CVPR}. The input is generally composed of an image and a textual query, with the former processed through the visual encoder and the projector, yielding fixed-length sequences of image features, which are then concatenated at the beginning of the textual feature input.
% \paragraph{Training Paradigm}
Pre-training primarily uses large-scale captioning instruction data, guiding the model to briefly describe images.
This phase enables the model to interpret image content, usually with LLMs' weights frozen and the visual encoder optionally updated. Some works such as Qwen-VL~\cite{bai2023qwen}, expand the pre-training to include additional tasks like visual question answering, updating the LLM component accordingly.
Supervised fine-tuning employs high-quality instruction data to enhance the LVLMs' ability to following diverse visual instructions and engaging in conversations. The visual encoder in this stage is typically kept static while the LLMs are tuned. During both stages, the projector is consistently updated, ensuring the model effectively bridges visual and textual data.

\section{Experimental Setup}
In this study, we conduct empirical experiments on Bunny-Llama-3-8B-V to investigate our hypothesis regarding the existence of a specific \textit{visual region} within LLMs (Sec.~\ref{section_region_position}$\sim$~\ref{section_region_scale_trend}), and apply our findings on LLaVA-1.5-7B, LLaVA-1.5-13B and Bunny-Phi3-mini-4B-V to validate its general applicability across different models (Sec.~\ref{section_validation}).

\subsection{LVLM Implementation}
% LLaVa-1.5, baseline: Vicuna-7B

We employ Bunny-Llama-3-8B-V for investigation, which builds upon the 32-layer Llama3-8B~\cite{touvron2023llama}, and LLaVA-1.5-7B/13B, built on the 32/40-layer Vicuna-1.5-7B/13B~\cite{vicuna2023}, Bunny-Phi3-mini-4B-V based on 32-layer Phi-3-mini for validation. Since the LLM components remain frozen during pre-training, we focus on the supervised fine-tuning stage using 695K and 665K language-image instruction-following instances for Bunny and LLaVA. Considering computational constraints, we use LoRA~\cite{hu2021lora}, highlighting that \textit{our approach is complementary to other efficient training methods}. 
% Training was conducted with DeepSpeed~\cite{song2023deepspeed4science} configured for zero3 optimization on 8×A800 GPUs. 
Additional implementation details are available in the Appendix.
% including a learning rate of 2e-4, rank of lora set to 128, and alpha of lora set to 256, while running the same number of training steps to ensure that updating gradient computations is maintained. We do not separately set the learning rate for the projector module. All models are implemented using Huggingface~\cite{wolf2019huggingface}, on NVIDIA GeForce RTX 4090 GPUs with 24 GB of memory, or NVIDIA A800 GPUs. Additionally, to expedite training on 4090 with limited GPU memory, we utilized DeepSpeed~\cite{song2023deepspeed4science} configured with zero2 instead of zero3, despite the resulting decrease in performance\footnote{Our experiments show that DeepSpeed with zero3 results in a fourfold increase in time consumption with performance improved by two points on average.}.

\subsection{Evaluation Tasks}
Our investigation spans 10 visual tasks involving both perception and cognition, to comprehensively evaluate models and examine our hypothesis.

\paragraph{Visual perception tasks} assess models' ability to interpret and understand surface-level visual features, like object identification and scene recognition, mirroring human sensory perception process. (1) OCRVQA~\cite{mishraICDAR19}: VQA by reading text in images through optical character recognition (OCR).  We follow\cite{bai2023qwen} for accuracy calculation on the test set, allowing a margin of error. (2) DocVQA~\cite{mathew2021docvqa}: VQA by interpreting document images. We use the same evaluation method and metric as OCRVQA on the validation set. (3) RefCOCOg~\cite{yu2016modeling}: A variant of RefCOCO~\cite{kazemzadeh2014referitgame} featuring more complex object referring expressions. We assess the reference expression generation on the test set using Intersection over Union metric. (4) TDIUC~\cite{kafle2017analysis}: covering 12 categories, primarily perception tasks (e.g., object presence, counting, recognition) with some cognition tasks (e.g., positional reasoning, affordance). Accuracy is measured on the validation set. 

\paragraph{Visual cognition tasks} require deeper reasoning based on visual stimuli, drawing on prior knowledge and decision-making abilities learned within LLMs, mirroring human cognitive thinking and manipulation.
\begin{table*}[!ht]
    \centering
    \setlength\tabcolsep{2pt}
    \resizebox{1.0\textwidth}{!}{
    \begin{tabular}{c|cccccccccc|c}
    \toprule
     % \multirow{2}{*}{Model Version} & \multicolumn{4}{c|}{Perception} & \multicolumn{6}{c|}{Cognition} & \multirow{2}{*}{Avg} \\
     %     & OCRVQA & DocVQA & RefCOCOg & TDIUC & MMBench & GQA & ScienceQA & TextVQA & MMMU & SEED-Imag\\
      Model Version & OCRVQA & DocVQA & RefCOCOg & TDIUC & MMBench & GQA & ScienceQA & TextVQA & MMMU & SEED-IMG & Avg \\
      \midrule
      All layers & 64.26\% & 29.45\% & 50.12\% & 83.84\% & 74.74\% & 64.29\% & 79.28\% & 62.11\% & 40.6\% & 73.13\% & 62.18\% \\
      \midrule
      \rowcolor[HTML]{E1E1E1} \multicolumn{12}{c}{\texttt{Heuristic Selections}} \\
      \midrule
      Sparse \& Uniform & 62.65\% & 29.51\% & \bf 48.33\% & 83.68\% & \bf 73.88\% & \bf 63.68\% & 78.78\% &  62.43\% & 42.1\%	& 72.61\% &	\bf 61.82\% \\
      Consecutive Lower & 61.38\% & 22.47\% & 46.49\%& 83.27\% & 73.63\% & 62.33\% & 75.26\% & 62.26\% & \bf 42.6\% & 72.66\% & 60.24\% \\
      Consecutive Lower-middle & 62.54\% & 26.13\% & 48.17\%& 83.77\% & 72.51\% & 62.81\% & 77.14\% & 60.96\% & 38.8\% & 72.16\% & 60.50\%  \\
      Consecutive Upper-middle & 62.32\% & 28.06\% & 43.12\% & 83.40\% & 70.27\% & 61.28\% & \bf 78.83\% & 59.33\% & 38.3\% & 70.45\% & 59.54\% \\
      Consecutive Top & 60.48\% & 26.47\% & 39.92\% & 83.22\% & 67.96\% & 60.30\% & 77.54\% & 58.71\% & 37.0\% & 71.00\% & 57.26\% \\

      Hybrid Top-Lower & 57.63\% & \bf29.76\% & 41.79\% & 83.26\% & 72.25\% & 62.71\% & 77.99\% & 62.74\% & 40.1\% & 72.59\% & 60.09\% \\
      \midrule
      \rowcolor[HTML]{E1E1E1} \multicolumn{12}{c}{\texttt{Importance-based Selections}} \\
      \midrule
      Image Attention Score & 63.65\% & 24.53\% & 43.62\% & \bf 83.90\% & 72.59\% & 62.82\% & 77.59\% & 61.99\% & 39.3\% & 72.29\% & 60.23\% \\
      Parameter Change Ratio & \bf 63.94\% & 26.94\% & 47.67\%& 83.88\% & 73.54\% & 63.21\% & 78.68\% & 61.73\% & 42.0\% & 72.85\% & 61.45\% \\
      Block Influence Score & 62.38\% & 28.45\% & 46.37\%& 83.73\% & 71.13\% & 61.93\% & 77.34\% & 59.93\% & 38.9\% & 71.66\% & 60.18\% \\
      Multimodal BI Score & 61.48\% & 28.80\% & 46.68\%& 83.74\% & 73.02\% & 63.23\% & 77.24\% & 62.23\% & 41.0\% & 72.25\% & 60.97\% \\
      Angular Distance & 60.95\% & 27.71\% & 46.74\%& 83.49\% & 73.88\% & 62.11\% & 77.14\% & \bf 62.76\% & 39.9\% & \bf 73.01\% & 60.77\% \\
    \bottomrule
    \end{tabular}
    }
    \caption{Performance comparison of Bunny-LLaMA-3-8B-V tuned with \textbf{\emph{different layer selection methods (8 layers)}}. Bold numbers indicate the best performance in each column (excluding ``all layers'').
    }
    \label{tab4:position_selection}
\end{table*}
(5) MMBench~\cite{liu2023mmbench}: focuses on cognition tasks, with some fine-grained perception tasks requiring knowledge and reasoning. For model variant comparison, we report accuracy on the dev subset instead of submitting to the evaluation server.
(6) GQA~\cite{hudson2019gqa}: real-world visual reasoning and compositional question answering.
(7) ScienceQA~\cite{lu2022learn}: sourced from elementary and high school science curricula, requiring external knowledge and reasoning. We evaluate only image-based questions.
(8) TextVQA~\cite{Singh_2019_CVPR}: requiring reasoning about text in images. 
(9) MMMU~\cite{yue2024mmmu}: covering math, science, and commonsense reasoning with accuracy calculated.
(10) SEED-IMG: The image-based QA from SEED-Bench~\cite{li2023seed}.
% For TextVQA, ScienceQA, and GQA, we use the LLaVA evaluation codes to measure accuracy.
    % \item NoCaps~\cite{agrawal2019nocaps}: The NoCaps dataset, standing for "novel object captioning at scale", consists of 166,100 human-generated captions describing 15,100 images from the Open Images validation and test sets. Sepcifically, we adopt its validation set and use the prompt "Please describe this image in one sentence". The CIDEr score~\cite{vedantam2015cider} is caculated to evaluate the performance of the model.  
    % \item VSR~\cite{liu2023VSR}: The VSR dataset is a benchmark for evaluating the spatial understanding of vision-language models, containing caption-image pairs with true or false labels, where each caption describes the spatial relation of two objects in the image. We utilize its test set and employ the prompt "Now I need you to help me judge whether the following description about location information is correct according to this picture. Just answer yes or no". The measure of performance is determined by the correctness of the responses.

% We use the Flickr30k~\cite{jia2015guiding} dataset to calculate the block influence score of each layer in the llava model backbone, the result of which is show

\section{Visual Region Investigation}
We first analyze the position and scale of the layerwise-structure vision region within its LLM core on Bunny-Llama-3-8B-V, to answer the following two questions. 

\subsection{Where are visual region layers located within LLMs for effective visual learning?}
\label{section_region_position}
To demonstrate the optimal positioning of the visual region in LLMs for effective and efficient visual learning, we re-train Bunny-Llama-3-8B-V by updating 25\% of layers (8 layers)~\footnote{We use the 8-layer configuration as a testbed for its balance of efficiency and effectiveness.} under various selection configurations. As pre-training does not involve LLM optimization, we focus on supervised fine-tuning, starting from the pre-trained checkpoint. We specifically explore different positional selection strategies as detailed below.
\begin{itemize}[itemsep=1pt, leftmargin=12pt]
    \item \textbf{Heuristic Layer Selection}
    (1) We intuitively hypothesize that tuning \textit{sparsely and uniformly distributed layers} (0,4,8,12,18,22,26,30) preserves LLMs' existing knowledge and reasoning abilities while enabling visual learning. (2) We experiment with tuning \textit{consecutive 8-layer blocks} at different positions in LLMs: lower layers (0$\sim$7), lower-middle layers (8$\sim$15), upper-middle layers (16$\sim$23), and top layers (24$\sim$31), with the latter being a common practice of \textbf{efficient domain-specific fine-tuning}~\cite{liao2024make}. (3) We test a hybrid of lower and top layers (0$\sim$3, 28$\sim$31).

    \item \textbf{Importance-based Layer Selection} We compare layer selection strategies based on varying importance metrics. (1) \textit{Image Attention Score}: We compute the average attention score on all image tokens at each layer to gauge the layer's affinity for image information. The top 8 layers with the highest scores are selected (1,2,3,4,5,27,29,31).
    (2) \textit{Parameter Change Ratio}~\cite{zhao2023unveiling}: 8 layers with the highest relative parameter change ratios (averaged all parameters in each layer) in Bunny-Llama-3-8B-V compared to its backbone Llama are selected (0,2,9,12,23,24,25,26). (3) \textit{Block Influence (BI) Score}~\cite{men2024shortgpt}: Using Flickr30k dataset, we calculate hidden state transformations at each layer as the BI score, and select 8 layers with the highest scores (12,15,18,25,27,29,30,31). (4) \textit{Multimodal BI Score}: We propose a multimodal variant that average  hidden state transformations respectively of visual tokens and textual tokens, and select 8 layers with highest scores (0,1,2,3,4,5,9,31). (5) \textit{Angular Distance Score}~\cite{gromov2024unreasonable}: The top 8 layers with the highest angular distances between consecutive layer inputs are selected (0,1,2,3,5,6,7,8). Detailed calculations for these metrics are provided in Appendix~\ref{appen:importance_metrics}.
\end{itemize}
\begin{table*}[!ht]
    \centering
    \setlength\tabcolsep{2pt}
    \resizebox{1.0\textwidth}{!}{
    \begin{tabular}{c|cccccccccc|c}
    \toprule
     % \multirow{2}{*}{Model Scale} & \multicolumn{4}{c|}{Perception} & \multicolumn{6}{c|}{Cognition} & \multirow{2}{*}{Avg} \\
     %     & OCRVQA & DocVQA & RefCOCOg & TDIUC & MMBench & GQA & ScienceQA & TextVQA & MMMU & SEED-Image\\
      Model Scale & OCRVQA & DocVQA & RefCOCOg & TDIUC & MMBench & GQA & ScienceQA & TextVQA & MMMU & SEED-IMG & Avg \\
      \midrule
      32 layers & \bf 64.26\% & 29.45\% & \bf 50.12\% & 83.84\% & 74.74\% & \bf 64.29\% & \bf 79.28\% & 62.11\% & 40.6\% & \bf 73.13\% & \bf 62.18\% \\
      16 layers & 62.42\% & 26.43\% & 49.15\% & \bf84.04\% & 74.83\% & 64.10\% & 78.93\% & \bf 62.96\% & \bf 42.6\% & 72.75\% & 61.82\%(99.42\%) \\
      8 layers & 62.65\% & 29.51\% & 48.33\% & 83.68\% & 73.88\% & 63.68\% & 78.78\% & 62.43\% & 42.1\% & 72.61\% & 61.78\%(99.36\%) \\
      6 layers & 62.25\% & \bf 29.76\% & 47.71\% &  84.01\% & \bf 75.00\% & 62.93\% & 77.54\% & 62.92\% & 40.6\% & 72.67\% & 61.55\%(98.99\%)\\
      4 layers & 62.40\% & 28.89\% & 46.00\% & 83.99\% & 73.71\% & 62.66\% & 77.69\% & 62.74\% & 39.2\% & 72.14\% & 60.94\%(98.01\%)\\
      2 layers & 57.96\% & 28.49\% & 44.67\% & 83.15\% & 72.68\% & 61.00\% & 78.48\% & 60.35\% & 40.8\% & 72.35\% & 60.00\%(96.49\%)\\
      1 layer & 53.68\% & 24.33\% & 38.47\% & 82.92\% & 68.64\% & 59.19\% & 77.69\% & 58.32\% & 37.4\% & 70.69\% & 57.14\%(91.89\%)\\
    \bottomrule
    \end{tabular}
    }
    \caption{Performance comparison of Bunny-Llama-3-8B-V fine-tuned with \textbf{\emph{different numbers of layers}}. Bold numbers represent the best performance in each column. Values in parentheses denotes the percentage relative to the performance achieved by tuning all layers.}
    \label{tab2:layer_number}
\end{table*}
\begin{figure*}[htbp]
    \centering
    \includegraphics[width=1.0\textwidth, height=4.5cm]{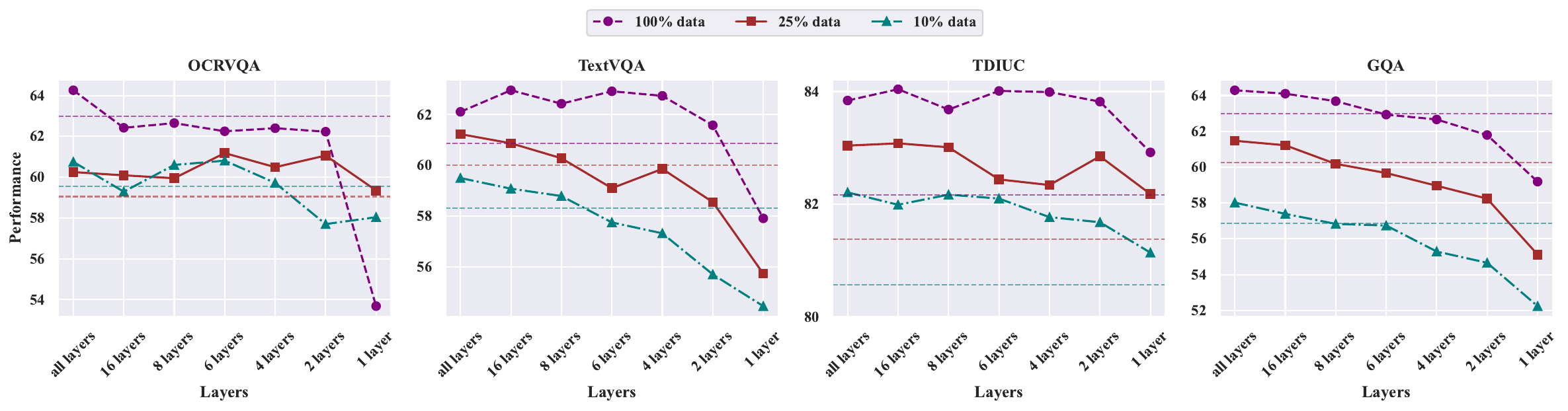}
    \caption{Performance variation of the re-trained Bunny-Llama-3-8B-V model across \textbf{\emph{different training data scales}} during the supervised fine-tuning stage, with tuning varying number of layers. Dashed lines indicate 98\% of the performance achieved by tuning all layers with the corresponding training data scale.}
    \label{fig:data_size}
\end{figure*}
The results are shown in Table~\ref{tab4:position_selection}. We observe that tuning sparsely and uniformly distributed layers achieves the best overall performance across perception and cognition tasks, closely matching the all-layers upper bound. In contrast, consecutive layers generally underperform, likely due to limited diversity in similar representations across adjacent layers~\cite{kornblith2019similarity}, which hinders adaptability to various tasks. This further underscores the superiority of sparsely and uniformly distributed layers. Notably, tuning top layers yields the worst performance, deviating from the conventional practice in domain-specific fine-tuning, where the last few layers are typically adjusted for downstream tasks~\cite{liao2024make}. This highlights a significant distinction between adapting to new modalities and new downstream domains. 

% Furthermore, while importance-based metrics are effective for layer pruning during LLMs inference, they are less effective than our empirically selected sparse and uniform layers for visual learning.
% Among importance-based selections, parameter change ratio serves as the best choice. 
While some importance-based selections, such as parameter change ratio, yield close performance, all importance-based methods operate post-hoc that require a fully trained model to compute importance metrics for layer selection. This makes them primarily suitable for inference and applying them during LVLM training incurs significantly higher computational costs. In contrast, our heuristic method is training-free, allowing for greater flexibility and direct transferability across different models, enhancing its practical applicability. We compare importance-based selections to show that our sparsely and uniformly distributed layers even outperform these post-hoc strategies and also simplify the process.

\subsection{What is the necessary scale of layers for effective and efficient LVLMs training?}
\label{section_region_scale}
To investigate the necessary scale of this visual region to enable LVLMs to receive visual signals and align with linguistic features, we re-train Bunny-Llama-3-8B-V by updating varying number of layers. We respectively experiment with configurations of 32, 16, 8, 6, 4, 2 and 1 layers, with all selected layers uniformly distributed across all layers~\footnote{Specifically, we select all even-numbered layers for the 16-layer configuration; layer 0, 4, 8, 12, 18, 22, 26, 30 for 8-layer; layer 0, 6, 12, 18, 24, 30 for 6-layer; and layer 0, 10, 20, 30 for 4-layer (Experiments show that layer 30 or 31 yields comparable results, and all odd-numbered selections perform slightly worse). 
% We also try replacing layer 30 with layer 31 which leads to a slight performance decline. 
Since 2-layer and 1-layer selection can not be uniform,
we have tested various configurations and adopted the best-performing strategy: layer 0 and 31 for 2-layer, and layer 31 for 1-layer based on highest block influence scores.}. 
This selection strategy is based on our finding that sparsely and uniformly distributed layers are the optimal position for effective visual learning. 
% We do not adopt prior pruning-based layer importance strategies for selecting tuned layers, such as block influence scores~\cite{men2024shortgpt} and layer similarity~\cite{gromov2024unreasonable}, since they are tailored for single modality within LLMs which are not suitable for visual learning.
% This can be illustrated in upcoming Sec.~\ref{section_4.3} where other layer selection methods show moderate performance. 

The results of tuning varying scales of layers on visual perception and cognition tasks are summarized in Table~\ref{tab2:layer_number}. Tuning 20$\sim$25\% of the layers (6 and 8 layers) retains approximately 98\% of the performance achieved by tuning all LLMs layers of Bunny-Llama-3-8B-V, with 25\% (8 layers) preserving up to 99\%. However, updating fewer than 4 layers leads to a significant performance drop, particularly in perception tasks that heavily relies on visual interpretation, highlighting the necessity of tuning at least 12.5\% of the layers (4 layers) for effective visual alignment.
% Tuning 6 layers can even outperform all layers across almost all datasets which suggests that tuning all LLMs layers may cause overfitting for visual-linguistic alignment or affect the LLMs' pre-existing knowledge and language reasoning abilities.

% \begin{figure*}[htbp]
%     \centering
%     \includegraphics[width=1.0\textwidth]{imgs/data_scale.pdf}
%     \caption{Performance variation of the re-trained Bunny-Llama-3-8B-V model across \textbf{\emph{different training data scales}} during the supervised fine-tuning stage, with tuning varying number of layers. Dashed lines indicate 98\% of the performance achieved by tuning all layers with the corresponding training data scale.}
%     \label{fig:data_size}
% \end{figure*}
\begin{table*}[!ht]
    \centering
    \setlength\tabcolsep{2pt}
    \resizebox{1.0\textwidth}{!}{
    \begin{tabular}{c|cccc|cccccc|c}
    \toprule
     % \multirow{2}{*}{Model Scale} & \multicolumn{4}{c|}{Perception} & \multicolumn{6}{c|}{Cognition} & \multirow{2}{*}{Avg} \\
     %     & OCRVQA & DocVQA & RefCOCOg & TDIUC & MMBench & GQA & ScienceQA & TextVQA & MMMU & SEED-Image\\
   Model Scale & OCRVQA & DocVQA & RefCOCOg & TDIUC & MMBench & GQA & ScienceQA & TextVQA & MMMU & SEED-IMG & Avg \\
    \midrule
      \multicolumn{12}{c}{LLaVA-1.5-7B} \\
      \midrule
      32 layers &  61.51\% & 19.46\% & 49.01\% & 83.40\% & 66.67\% & 62.98\% & 68.47\% & 58.19\% & 35.3\% & 67.52\% & 57.25\% \\
      16 layers & 64.01\% & 20.75\% & 48.02\% & 83.47\% & 64.00\% & 62.43\% & 67.53\% & 58.27\% & 35.4\% & 67.22\% & 57.11\%(99.76\%) \\
      8 layers & 62.19\% & 21.10\% & 47.71\% & 83.10\% & 63.92\% & 61.60\% & 68.17\% & 57.35\% & 34.6\% & 67.23\% & 56.70\%(99.04\%) \\
      6 layers & 61.39\% & 22.84\% & 46.54\% & 83.31\% & 61.77\% & 61.08\% & 68.32\% & 56.19\% & 33.2\% & 65.69\% & 56.04\%(97.87\%) \\
      4 layers & 63.28\% & 21.01\% & 43.47\% & 83.14\% & 60.82\% & 60.48\% & 67.97\% & 54.48\% & 33.8\% & 64.08\% & 55.25\%(96.51\%) \\
      2 layers & 54.54\% & 19.10\% & 41.90\% & 81.47\% & 57.22\% & 57.38\% & 65.84\% & 53.27\%  & 33.7\% & 63.19\% & 52.76\%(92.16\%)\\
      1 layer & 53.16\% & 16.96\% & 33.29\% & 81.20\% & 51.89\% & 55.83\% & 64.50\% & 45.51\% & 30.1\% & 57.64\% & 49.01\%(85.61\%) \\
      \midrule
      \multicolumn{12}{c}{LLaVA-1.5-13B} \\
      \midrule
      40 layers & 67.60\% & 25.19\% & 50.26\% & 83.61\% & 68.38\% & 63.29\% & 71.64\% & 60.21\% & 37.2\% & 68.70\% & 59.61\% \\
      10 layers & 65.17\% & 23.56\% & 48.27\% & 83.57\% & 66.58\% & 62.01\% & 70.75\% & 59.13\% & 36.9\% & 67.39\% & 58.33\%(97.85\%) \\
      9 layers & 66.47\% & 23.65\% & 49.29\% & 83.74\% & 65.61\% & 62.31\% & 72.14\% & 59.71\% & 37.7\% & 67.29\% & 58.80\%(98.64\%) \\
      \midrule
      \multicolumn{12}{c}{Bunny-Phi3-mini-4B-V} \\
      \midrule
      32 layers & 63.62\% & 29.19\% & 48.07\% & 83.69\% & 72.94\% & 62.35\% & 76.75\% & 60.64\% & 42.4\% & 72.09\% & 61.17\% \\
      8 layers & 61.96\% & 27.21\% & 46.95\% & 83.11\% & 71.74\% & 61.38\% & 75.71\% & 59.69\% & 42.3\% & 71.53\% & 60.16\%(98.35\%)\\
    \bottomrule
    \end{tabular}
    }
    \caption{Performance of LVLMs with varying LLM backbones and parameter scales tuned with different numbers of layers. Values in parentheses denotes the percentage relative to the performance achieved by tuning all layers.}
    \label{tab:llava_layer_number}
\end{table*}
\subsection{Trend between Data Size and Visual Region Scale} 
\label{section_region_scale_trend}
We further explore the trend between data size and the optimal layer count for effective visual instruction tuning. Using random subsets of 100\%, 25\% and 10\% from a pool of 695K visual instruction-following instances, we tune Bunny-Llama-3-8B-V with varying numbers of layers following the same selection strategy as the full dataset. We report the performance trends across four datasets, OCRVQA, TextVQA, TDIUC and GQA. 
As shown in Figure~\ref{fig:data_size}, tuning 25\% of the layers consistently achieves over 98\% of full performance across different data sizes while reducing training time. This approach offers a resource-efficient pathway for optimizing hyperparameters and training data selection by tuning such a visual region before finalizing the model with all layers. Moreover, even with smaller datasets, tuning fewer than 4 layers still results in notable performance declines. 

\section{Further Analysis}
\subsection{Generalizability Validation}
\label{section_validation}
% Our approach extends seamlessly to other LVLMs. By adjusting the proportion of trainable layers and leveraging LoRA-based fine-tuning, the methodology balances computational efficiency and performance across models with varying architectures and sizes. This adaptability highlights its broad applicability to a range of LVLMs.
To validate our findings of the visual region beyond Bunny-Llama-3-8B-V, we take LLaVA-1.5-7B, LLaVA-1.5-13B and Bunny-Phi3-mini-4B-V as additional testbeds to assess the generalizability across LVLMs with different LLM backbones and parameter scales. Following the setup in Sec.~\ref{section_region_scale}, we re-train these models with different number of layers that are sparsely and uniformly distributed within their respective backbones, including Vicuna-1.5-7B, Vicuna-1.5-13B and Phi-3-mini-4B~\cite{abdin2024phi}. Results presented in Table~\ref{tab:llava_layer_number} show that under our visual region positioning strategy, tuning approximately 25\% of the layers consistently yield 98\% of the full performance. This demonstrates that our approach generalizes effectively across varying LVLMs.

\subsection{Computational Cost}
\begin{figure}[!th]
    \centering
    \includegraphics[width=0.46\textwidth, height=5cm]{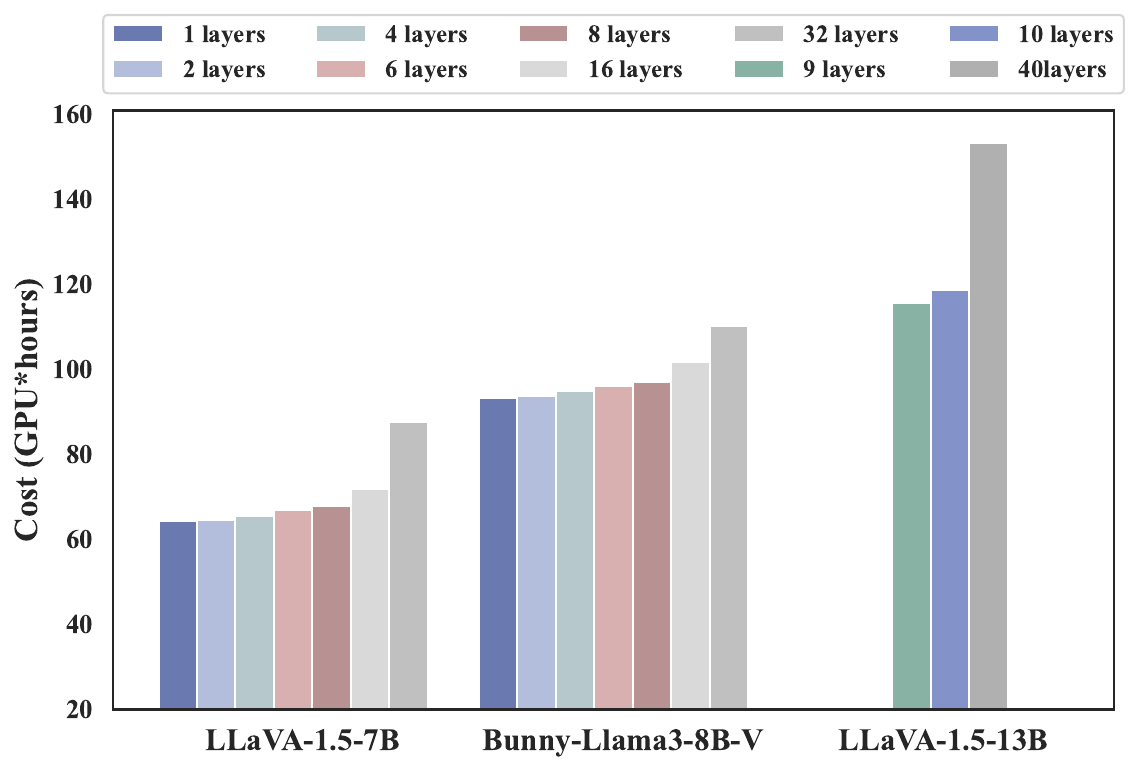}
    \caption{Computational costs for tuning LLaVA-1.5-7B, Bunny-Llama-3-8B-V, and LLaVA-1.5-13B with different number of layers using LoRA.}
    \label{fig:computational_cost}
\end{figure}
To demonstrate the efficiency of visual region-based tuning, we report the computational costs associated with tuning different numbers of layers across various models using the LoRA strategy. For fair comparison across setups with different numbers of GPUs (specifically A800 GPUs in this analysis), we compute the product of the number of GPUs and running hours as a measure of computational cost. 
From Figure~\ref{fig:computational_cost}, Table~\ref{tab2:layer_number} and Table~\ref{tab:llava_layer_number}, 
tuning a visual region comprising up to 25\% of layers (8 layers for LLaVA-1.5-7B and Bunny-Llama3-8B-V, 10 layers for LLaVA-1.5-13B) can achieve 98\% of full performance while achieving substantial reductions in computational overhead. Specifically, we reduce training time by 23\% for LLaVA models and 13\% for Bunny. These results highlight that the effectiveness of visual region-based tuning in training LVLMs efficiently with minimal performance trade-offs.
Moreover, this relative reduction in computational cost would be more significant as dataset and model sizes scale.

\subsection{Evaluation of Textual Tasks}
As highlighted in~\cite{dai2024nvlm, agrawal2024pixtral} and illustrated in Figure~\ref{fig:layer_drop}, multimodal training risks degradation of LLMs' pre-trained linguistic knowledge and reasoning capabilities. 
To verify whether training our sparsely and uniformly distributed visual region affects the model linguistic capacity, we extend our evaluation to four text-only question answering datasets, MMLU~\cite{hendrycks2020measuring}, C-Eval~\cite{huang2023c}, CMMLU~\cite{li2023cmmlu}, and BIG-bench-Hard~\cite{suzgun2022challenging}, covering diverse topics and fields. We use ``Answer with the option's letter from the given choices directly'' as the prompts for the first three and ``Please answer this question in a word or phrase'' for BIG-bench-Hard, and allow models to provide explanations alongside its responses. We adopt a five-shot prompting strategy for MMLU, C-Eval and CMMLU, and a zero-shot strategy for BIG-bench-Hard.
% We calculate the multi-choice accuracy as the evaluation metric, allowing models to provide additional explanations alongside its responses. 
\begin{table}[th!]
    \centering
    \setlength\tabcolsep{2.5pt}
    \resizebox{0.49\textwidth}{!}{
    \begin{tabular}{c|cccc}
    \toprule
     Model Version & MMLU & BIG-Bench-H & C-Eval & CMMLU \\
      \midrule
      \multicolumn{5}{c}{Bunny-LLaMA3-8B-V} \\
      \midrule
        Fully-trained (32layers) & 60.27\% & 30.93\% & 45.84\% & 45.68\%\\
        Partial-trained (8layers) & 63.36\% & 31.50\% & 49.70\% & 48.39\%\\
        LLM-Backbone & 66.01\% & 57.93\% & 50.52\% & 50.70\% \\
        \midrule
      \multicolumn{5}{c}{LLaVA-1.5-7B} \\
      \midrule
        Fully-trained (32layers) & 50.52\% & 26.85\% & 38.34\% & 37.27\% \\
        Partial-trained (8layers) & 50.74\% & 31.64\% & 39.08\% & 37.71\% \\
        LLM-Backbone & 49.78\% & 29.33\% & 38.78\% & 36.60\% \\
    \bottomrule
    \end{tabular}
}
    \caption{Performance on text-only tasks. The LLm backbones of Bunny-LLaMA3-8B-V and LLaVA-1.5-7B are respectively LLaMA3-8B and Vicuna-1.5-7B. }
    \label{tab7:textual_eval}
\end{table}

As shown in Table~\ref{tab7:textual_eval}, fully-trained LVLMs generally exhibit decreased performance on text-only tasks compared to their LLM backbones, particularly with more powerful LLaMA3-8B and on the challenging BIG-bench-Hard dataset. In contrast, our selectively trained LVLMs 
minimally compromise models' linguistic capacity, which 
consistently outperform fully-trained LVLMs, and sometimes even surpass their LLMs backbones. These results support our hypothesis that positioning the visual region strategically by tuning sparsely and uniformly distributed layers better preserves LLMs' linguistic knowledge and reasoning capabilities, whereas full training may cause minor disruptions.

\section{Visual Region-Based Layer Pruning}
Beyond layer selection for efficient LVLMs training, we explore whether the visual region can also benefit LVLM efficient inference. 
Although layer pruning techniques~\cite{men2024shortgpt, ma2023llm} have been widely developed for LLM inference, they prove ineffective for LVLMs. As shown in Figure~\ref{fig:layer_drop} (right), minimal layer removal causing significant performance degradation on visual tasks even using advanced angular distance based pruning strategy~\cite{gromov2024unreasonable}.
\begin{figure}[!ht]
    \centering
    \includegraphics[width=0.5\textwidth]{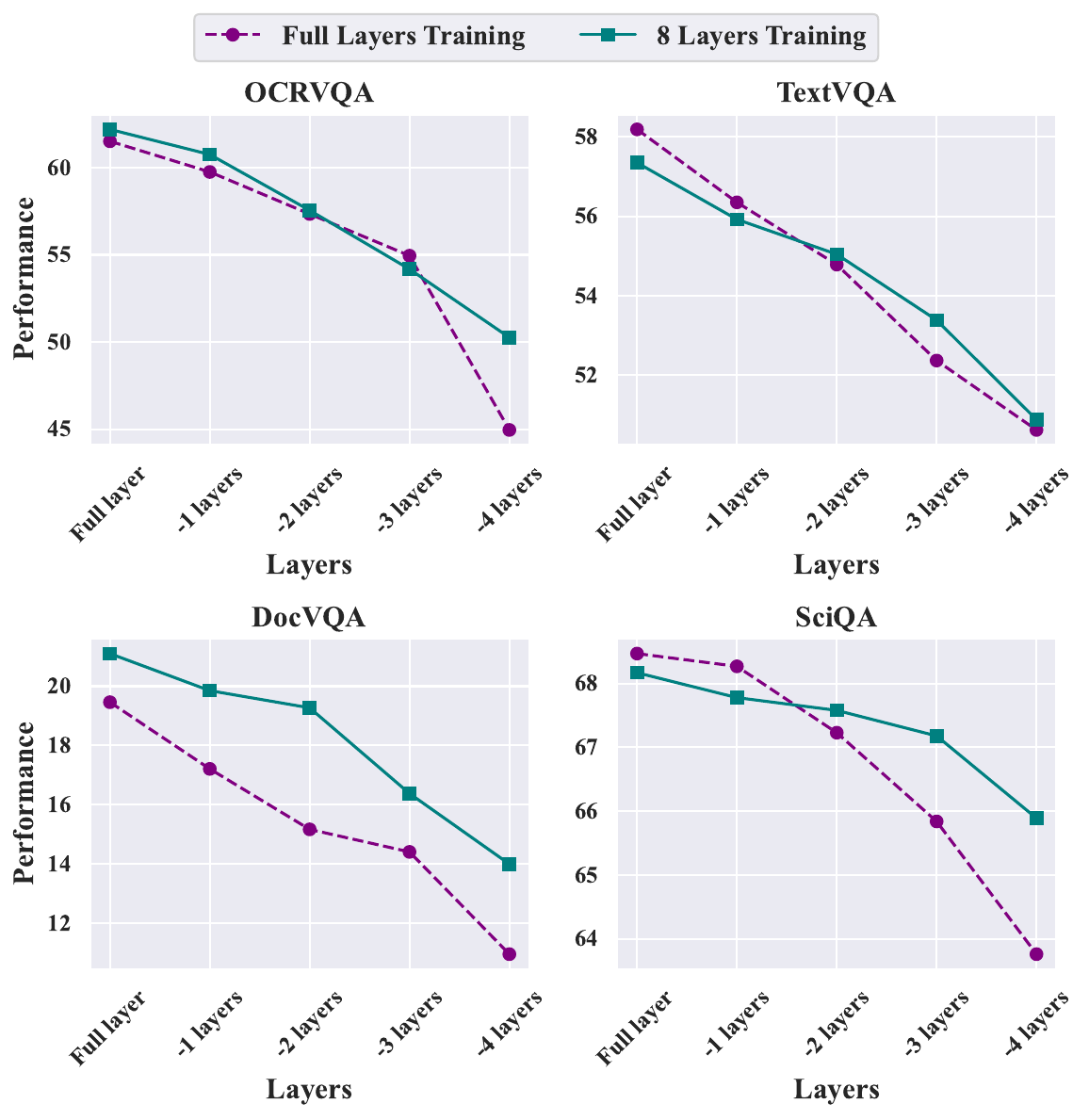}
    \caption{Results of pruning LLaVA-1.5-7B using angular distance-based strategy with 0$\sim$4 layers removed. Dashed lines represent pruning applied to the fully trained model while solid layers denote our visual region-based pruning within the targeted trained model.}
    \label{fig:layer_pruning}
\end{figure}

Building on our visual region targeted training, we propose a visual region-based pruning paradigm that selectively prunes less-important layers outside the visual region after training. Specifically, we follow the angular distance based layer importance metric and select 0$\sim$4 layers with the lowest angular distance outside the visual region. We do not evaluate pruning beyond this range as removing additional layers in LVLMs would lead to significant performance collapse. We evaluate this approach on LLaVA-1.5-7B across four datasets: OCRVQA, TextVQA, DocVQA and SciQA. As shown in Figure~\ref{fig:layer_pruning}, our paradigm generally maintain higher performance, especially when pruning 3$\sim$4 layers, even though the visual region targeted trained model performs slightly worse than fully trained model without pruning. This result demonstrates that our paradigm effectively minimizes performance degradation compared to pruning in full-layer trained LVLMs, serving as an initial exploration into LVLM-specific pruning strategies.

\section{Related Work}
% \subsection{Visual Instruction Tuning}
% Recent research community has witnessed an emergent interest in LVLMs~\cite{li2023blip,zhu2023minigpt,bai2023qwen,liu2024visual} due to their unparalleled ability to interpret and interact with the world via both visual and linguistic channels. These models demonstrate impressive performance across diverse visual tasks~\cite{lu2022learn, liu2023mmbench, fu2024mme}, powered by a architecture that typically integrates a visual encoder for image feature extraction~\cite{radford2021learning}, a projector or connection module to align these features with linguistic input~\cite{tsimpoukelli2021multimodal, alayrac2022flamingo}, and a LLM~\cite{chung2022scaling, vicuna2023} serving as a knowledge and reasoning foundation. LVLMs undergo training through a process called visual instruction tuning~\cite{liu2023visual}, which is basically divided into pre-training and fine-tuning stages. 
% They leverages large-scale visual instruction-following datasets, enabling LLMs to generalize to diverse visual tasks. These datasets, often reformed from existing collections of image-text pairs, including tasks like object detection, image captioning, and visual question answering~\cite{sharma2018conceptual, laioncoco, changpinyo2021conceptual, hudson2019gqa, Singh_2019_CVPR}. This process involves both training the projector and LLMs, with the visual encoder's weights being occasionally updated to further enhance performance~\cite{bai2023qwen}.

\subsection{Efficient Training and Inference}
Recent research community has witnessed an emergent interest in LLMs~\cite{touvron2023llama, vicuna2023} and LVLMs~\cite{li2023blip,zhu2023minigpt,bai2023qwen,liu2024visual} due to their remarkable ability to interpret and interact with the world via linguistic and visual channels. With the sustainably increased scale of LLMs and LVLMs, training or inference using all model parameters are cost for practical deployment. There are numerous techniques for efficient model training and inference. For instance, quantization reduce the memory footprint of models by decreasing the precision of model weights~\cite{dettmers2208llm, dettmers2023case, xiao2023smoothquant}. Low rank adapters enable cost-effective fine-tuning by updating only a small subset of the adapter parameters~\cite{hu2021lora, karimi2021compacter}.

Moreover, LLMs exhibit significant redundancy at the layer level, making training or inference with all layers computationally wasteful, and this redundancy is established for LVLMs as well, where LLMs serve as the core cognitive brain for visual learning.
In responding, layer-wise freezing techniques~\cite{zhang2024unified, liang2023less, pan2024lisa} and layer pruning strategies~\cite{men2024shortgpt, ma2023llm, gromov2024unreasonable} are proposed to enable efficient LLM fine-tuning and inference.
However, they are designed for LLMs and fail to generalize effectively to visual learning, often resulting in substantial performance degradation.
While~\citet{zhang2024overcoming} introduce parameter localization for visual tasks, their approach is highly task-specific and data-dependent, limiting its applicability to versatile visual learning and neglecting the preservation of linguistic capabilities.
In contrast, we propose a more efficient layer-selected strategy for LVLMs training and inference. 

\subsection{Functional Regions in LLMs}
The existing literature on cognitive science and brain localization indicates that different regions among the human brain are dedicated to specific functions~\cite{fedorenko2016language}, such as frontotemporal language processing region localized by
~\citet{scott2017new}. ~\citet{grill2004human} highlight the existence of visual regions in neuroscience~\cite{grill2004human}. These insights have inspired an analogy with LLMs, increasingly viewed as cognitive core for remarkable performance across diverse tasks, mirroring the human brain's functionality in terms of overall planning and processing. For example, ~\citet{aw2023instruction} propose that LLMs can be aligned to the human brain through instruction-tuning.
Building upon this parallel, ~\citet{zhao2023unveiling} unveil a core linguistic region within LLMs, accounting approximately 1\% of the model's parameters. ~\citet{li2024memory} identify a duality between Tulving’s synergistic ecphory model (SEM) of memory and LLMs' emergent abilities. Drawing inspiration from these, our research focuses on defining a vision region within LLMs, suggesting a more effective and efficient pipeline to optimizing LVLMs for visual tasks.

\section{Conclusion}
In this study, we introduce an effective and efficient training paradigm for LVLMs by activating a specific \textit{visual region} within LLMs. This offers a new pipeline for advancing LVLMs which first identify such \textit{visual region} using limited data followed by efficient continual training. Specifically, we investigating the necessity of tuning all layers within LLM cores, and propose the concept of a specialized \textit{visual region} within LLMs. We conduct extensive empirical experiments with Bunny-LLaMA-3-8B-V, covering a range of visual and textual tasks. Our results reveal that selectively updating no more than 25\% of sparsely and uniformly layers, can preserve nearly 99\% visual performance, while also yielding comparable results in textual tasks. This targeted LVLMs' training approach is consistently effective for different models and parameter scales, effectively reducing training time by 23\% for LLaVA models and 12\% for Bunny-LLaMA-3-8B-V. Additionally, we propose a visual region-based layer pruning by strategy removing non-critical layers outside the visual region and achieve minimal performance drop. Overall, our work presents a promising pathway for more efficient LVLMs training and inference, while complementing existing efficient training methods.

% Considering the notable improvements observed in visual perception tasks, we introduce a targeted training strategy focusing on these \textit{visual region} layers for visual perception learning, with all layers dedicated to cognition tuning, leading to general superior performance. 

\section*{Limitations}
\paragraph{Experimented Models} Our work primarily focuses on LLaVA-1.5 family, Bunny-LLama3-8B-V and Bunny-Phi3-mini-4B-V to demonstrate the effectiveness and efficiency of our proposed training and inference paradigms for LVLMs.
Future work will expand to a broader range of models to further validate the generalizability of our approach. Additionally, we will explore extensions to other modalities such as speech, and investigate the existence of other modality-specific regions to develop more versatile and scalable multimodal models.

% \paragraph{Extend to models with more modality} While our work primarily focuses on visual-language models (VLMs) by activating a specific \textit{visual region} within LLMs, future research should explore the extension of this paradigm to models incorporating additional modalities, such as audio, speech, or other modality inputs. This will allow us to investigate whether the concept of a modality-specific region in LLMs, akin to the visual region, can be generalized to effectively build more versatile and scalable multimodal models that can more efficiently understand and generate information of any modalities simultaneously. 

% \paragraph{Extend to models with higher computational efficiency}  Although our approach can effectively save training costs by activating \textit{visual region}, activating the whole model parameters during inference decrease the computational efficiency, In the future, we will integrate the method of activating the \textit{visual region} with sparse model architectures. Explore a routing mechanism targeting modality-specific partitions within the model to implement a sparse model with modality-specific functional partitions, similar to the functional regions of the human brain with higher computational efficiency both in training and inference.
\paragraph{Sparse Architectures} While our approach effectively reduces training and inference costs by activating the \textit{visual region}, it currently operate in a layer-wise dense manner. Future efforts will focus on integrating our method with sparse model architectures to optimize \textit{visual region} activation. For example, explore routing mechanisms targeting modality-specific partitions within models to implement sparse mixture-of-expert architectures with specialized functional areas, analogous to the functional regions of the human brain.

% Bibliography entries for the entire Anthology, followed by custom entries
%\bibliography{anthology,custom}
% Custom bibliography entries only
\bibliography{custom}

\appendix

\section{Details of Layer Importance Metrics}
\label{appen:importance_metrics}
\begin{figure*}[h]
    \centering
    \includegraphics[width=1\textwidth]{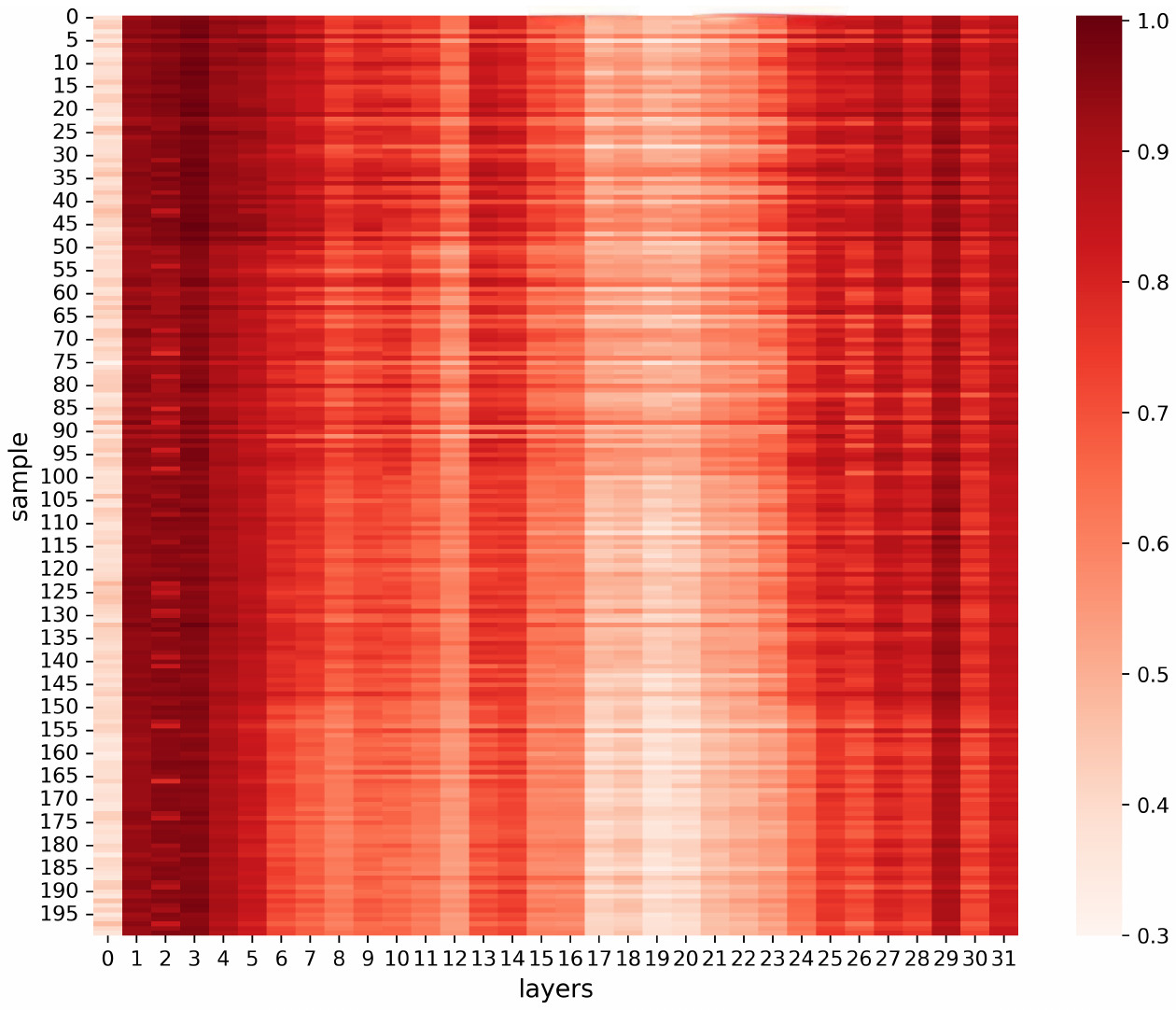}
    \caption{Visualization of Image Attention Scores for every instances across all layers}
    \label{fig:attention score map}
\end{figure*}

To demonstrate the effectiveness of our heuristically identified sparsely and uniformly distributed visual region, we conduct a comparative analysis against several other layer importance metrics (originally for layer pruning) by selecting 8 layers and re-training Bunny-Llama-3-8B-V. Below are the details of how these metrics are calculated. 
\begin{itemize}[itemsep=1pt, leftmargin=12pt]
    \item \textbf{Block Influence (BI) Score~\cite{men2024shortgpt}}: serves as an indicator of layer importance by measuring the transformation of hidden states. We utilize the Flickr30k dataset~\cite{jia2015guiding} to calculate the BI score for each layer within LVLMs. The BI score of $i^{th}$ layers is calculated as following:
    \begin{align}
    BI_{i}=1-\mathbb{E}_{X,t}\frac{X^{T}_{i,t}X_{i+1,t}}{\lVert X_{i}\rVert_2 \lVert X_{i+1}\rVert_2} \nonumber
    \end{align}
    where $X_{i}$ represents the hidden states of the $i^{th}$ layer and $X_{i,t}$ denotes the hidden states of the $t^{th}$ token at the $i^{th}$ layer. By calculating the average cosine similarity of token states before and after passing through a layer, we measure the change magnitude across all tokens.
    \item \textbf{Multimodal BI Score}: As the above method treats visual image and text as a single modality, we propose a multimodal variant that separately calculates the hidden state transformations of visual tokens and textual tokens, and take its average as a multimodal BI score. The Multimodal BI score of $i^{th}$ layers is calculated as follows.
    \begin{align}
    BI'_{i}=1-\frac{1}{2}(\mathbb{E}_{X,t}\frac{X^{T}_{i,t}X_{i+1,t}}{\lVert X_{i}\rVert_2 \lVert X_{i+1}\rVert_2} \nonumber \\
    +\mathbb{E}_{Y,l}\frac{Y^{T}_{i,l}Y_{i+1,l}}{\lVert Y_{i}\rVert_2 \lVert Y_{i+1}\rVert_2})    \nonumber 
    \end{align}
    $X_{i,t}$ and $Y_{i,l}$ respectively mean the hidden states of the $t^{th}$ visual token and the $l^{th}$ text token at the $i^{th}$ layer. We calculate the cosine similarity of each modality tokens before and after passing through a layer, then average the results. This balances the token quantity across various modalities.
    \item \textbf{Parameter Change Ratio~\cite{zhao2023unveiling}}: We calculate the relative change ratio of the parameters in LVLM against its backbone LLM across each layer (by averaging all parameters within each layer). The parameter change ratio of $i^{th}$ layers is calculated as follows:
    \begin{align}
    R_{i}=\mathbb{E}_{\theta\in L_{i},j}\lvert\frac{\theta'_{j}-\theta_{j}}{\theta_{j}}\rvert \nonumber
    \end{align}
    where $\theta_{j}$ and $\theta'_{j}$ respectively mean the $j^{th}$ parameter of layer $L_{i}$ in LLM and LVLM.
    \item \textbf{Angular Distance~\cite{gromov2024unreasonable}}: We calculate the Angular Distance of the parameters in LVLM against its backbone LLM across each layer (by averaging all parameters within each layer). The Angular Distance of $i^{th}$ layers is calculated as follows:
    \begin{align}
   D_{i} = \frac{1}{\pi} \arccos \left( \frac{\theta'_{j} \cdot \theta_{j}}{\|\theta'_{j}\| \|\theta_{j}\|} \right)
   \nonumber
    \end{align}
    where $\theta_{j}$ and $\theta'_{j}$ respectively mean the $j^{th}$ parameter of layer $L_{i}$ in LLM and LVLM, $ \|\cdot\|\ $ denotes the $L^2$-norm and the factor of $\frac{1}{\pi}$ is a constant. 
    \item \textbf{Image Attention Score}: We calculate image attention score to measure each layer's affinity for image information. We utilize the DocVQA, OCRVQA, TDIUC, and RefCOCOg datasets, sampling 50 instances from each dataset to calculate the attention scores of the all image tokens for each layer within Bunny-Llama-3-8B-V. The heat map of image attention Score of every instances for each layers in Bunny-Llama-3-8B-V is showed in Figure~\ref{fig:attention score map}. The image attention score of one instance in $i^{th}$ layers $A_i$ is calculated as follows:
    \begin{align}
    A_i = \frac{\sum_{t=\text{k}}^{\text{k} + N_{\text{img}} - 1} \sum_{h=1}^H \sum_{j=1}^T\text{Attn}[i][h, j, t] }{N_{\text{img}} H} 
    \nonumber
    \end{align}
    where $H$ represents the number of attention heads per layer and $T$ denotes the total number of tokens at the $i^{th}$ layer. $N_\text{img}$ is the number of image tokens of the instance. The index range for the image tokens is from $k$ to $k + N_\text{img} -1$. While $\text{Attn}[h, j, t]$ means the attention score of the $h^{th}$ attention head for the $j^{th}$ token to the $t^{th}$ token.
    
\end{itemize}

\end{document}